# Probabilistic Neural Network with Complex Exponential Activation Functions in Image Recognition using Deep Learning Framework

Andrey V. Savchenko

*Abstract*— **If the training dataset is not very large, image recognition is usually implemented with the transfer learning methods. In these methods the features are extracted using a deep convolutional neural network, which was preliminarily trained with an external very-large dataset. In this paper we consider the nonparametric classification of extracted feature vectors with the probabilistic neural network (PNN). The number of neurons at the pattern layer of the PNN is equal to the database size, which causes the low recognition performance and high memory space complexity of this network. We propose to overcome these drawbacks by replacing the exponential activation function in the Gaussian Parzen kernel to the complex exponential functions in the Fejér kernel. We demonstrate that in this case it is possible to implement the network with the number of neurons in the pattern layer proportional to the cubic root of the database size. Thus, the proposed modification of the PNN makes it possible to significantly decrease runtime and memory complexities without loosing its main advantages, namely, extremely fast training procedure and the convergence to the optimal Bayesian decision. An experimental study in visual object category classification and unconstrained face recognition with contemporary deep neural networks have shown, that our approach obtains very efficient and rather accurate decisions for the small training sample in comparison with the well-known classifiers.**

## I. Introduction

IMAGE recognition tasks nowadays are usually solved with contemporary deep neural networks [1], [2]. Such neural network models as convolutional neural networks (CNN) have reached a certain level of maturity when the size of the training set is large with respect to the number of different classes [3]. For instance, these models proved to be very efficient in such tasks, as optical character recognition [4], [5],

This work was conducted in the Laboratory of Algorithms and Technologies for Network Analysis, National Research University Higher School of Economics and was supported by RSF (Russian Science Foundation) project No. 14-41-00039.

A.V. Savchenko is with the Laboratory of Algorithms and Technologies for Network Analysis, National Research University Higher School of Economics, 25/12 Bolshaya Pecherskaya St., Nizhny Novgorod 603155, Russia (e-mail: avsavchenko@hse.ru).

visual object category recognition [6]–[9], pedestrian identification [10], traffic sign classification [11], etc. The task becomes much more complicated if the training database is not very large and contains only a small number of images per each class [12], [13]. In such a case, the problem becomes very challenging due to the known variability of images and conditions of observations, e.g., illumination, noise, etc. [2]. To solve the image recognition task, an observed image and all instances in the database are processed in order to extract appropriate visual features [14]. A decade ago the traditional features of computer vision [2], e.g., SIFT [15], SURF [16], or HOG [17], [18], were typically used. Now it is more common to apply the transfer learning techniques [1], [19]. The deep CNN is trained with a very large dataset, and the output of one of the last layers of this network for the input image is used as its feature vector. After such extraction of feature vectors, any appropriate classifier can be used to make a final decision [20].

If the size of the training sample is relatively small, the error rate of such state-of-the-art classification methods as support vector machines (SVM) [20], [21] and the multi-layer perceptrons (MLP or feed-forward neural networks) [22], is usually too high [23], [24]. This small sample size problem is crucial in several recognition tasks, e.g., in the face recognition when only one reference image of each identity is sometimes available [25], [26]. In such a case the instance-based learning approach [27] can be applied, e.g. the k-nearest neighbor (k-NN) method [20], [28]. A variant of this approach is implemented in the DeepFace method [29], which solves the face verification task [30] by training a CNN using an external large face database. The resulted high-dimensional feature vectors were classified using the 1-NN method with an appropriate dissimilarity measure (Euclidean metric, chi-squared distance, etc.) [29], [31].

Another possible solution for such a problem is the application of statistical approach [32], in which the image recognition is reduced to a testing of hypothesis about distribution of an observed feature vector. The unknown probability density functions of each class (likelihoods) are estimated using either parametric or nonparametric techniques [20]. According to the parametric approach, the probability distribution family is fixed, and the parameters are estimated based on the given training sample. The most widely used family is the Gaussian distribution of the feature vectors or the components extracted with the PCA (Principal Components Analysis) [2], [33]. One major limitation of this approach is that it works well only when the assumptions about the type of distribution are satisfied [34]. The effectiveness depends on the various conditions under which the models are developed and should be thoroughly tested before the models can be successfully applied. Thus, the data driven self-adaptive nonparametric methods have been widely used in practice [35], [36]. These methods adjust themselves to the data without any explicit specification [34]. Such techniques are based on nonlinear models, which make them flexible in modeling real world complex relationships and were proved to converge to the Bayes-optimal decision surface.

The well-known parallel implementation of the nonparametric approach is the probabilistic neural network (PNN) [37], [38]. D. Specht called his method a neural network because of its natural mapping onto a feedforward network with four layers, in which an exponential function is used instead of the sigmoid activation function [34]. The PNN approximates the class-conditional probability distributions by the sum of identical isotropic Gaussians [39]. In practice, the PNN is characterized by extremely fast training procedure with possibility to add new instances and classes in real time [12]. In fact, it is often an excellent pattern classifier for the small sample size problem, outperforming other classifiers including back propagation. The various modifications of the PNN were applied in such real-world applications, as face recognition [18], image-based handwritten signature verification [40], authorship attribution [41], 3D objects and handwritten digits recognition [42], speech recognition [12], [43], medical data classification [44], defect recognition [45], etc.

However, there are well-known disadvantages of the PNN caused by the implementation of the instance-based learning in this classifier [27]. First of all, its time complexity can be very low, because the number of neurons in the pattern layer is equal to the number of instances in the database [38]. Hence, the network is based on an exhaustive search through all training sample [28]. Secondly, it requires large memory to store all training samples (so called memory-based approach to classification). Finally, the accuracy of the PNN significantly depends on the proper choice of the smoothing parameter [41]. There are several papers, which try to deal with the mentioned drawbacks. For instance, the need for anisotropic Gaussians for each class was clear right after the introduction of the PNN [39], but the procedures to obtain proper estimates of many smoothing parameters usually require rather large training set. Nowadays the smoothing parameter is optimized with the differential evolution algorithm [45] or reinforcement learning [46]. It is also known, that the performance of the PNN can be improved by selecting the most typical instances (prototypes) by using the clustering of the training set [27], [44]. Unfortunately, in this case the decision is not optimal in the Bayesian sense, especially, if the clusters are imbalanced. Moreover, the training procedure becomes time-consuming [20], so the main advantage of this neural network does not hold anymore.

Thus, in this paper we examine an alternative possibility to increase the speed [14] of image classification using the deep learning framework [3] and decrease the memory space complexity of the PNN while preserving its Bayes-optimality property. Namely, we propose the novel modification of the PNN, in which the Gaussian activation function is replaced to the complex exponential function, i.e., the Parzen kernel is replaced to the Fejér kernel and the densities are estimated with the orthogonal Fourier series [47]–[49]. It is known that the orthogonal series density estimates converge if the number of terms (cut-off) is proportional to the cubic root of the number of observations [50]. Based on these results, we present an efficient implementation of the PNN with the Fejér kernel. Our implementation does not require the instance-based learning, so it is much faster, than the original PNN. It is demonstrated that the modified network still saves all advantages of the original PNN, such as fast training procedure and convergence to the Bayes decision.

The rest of the paper is organized as follows. Section 2 presents an application of the PNN [38] in image recognition using features extracted with the deep CNN [3]. In Section 3, our modification of the PNN with the complex exponential functions is introduced. In Section 4 we present the experimental results in visual object category recognition [51] and unconstrained face recognition [52], [53] using the well-known datasets (Caltech101, Caltech256, Stanford Dogs, PubFig83 and CASIA-WebFace) and the feature extraction with the popular deep neural networks architectures, e.g., GoogLeNet [8] and VGGNet [7]. Finally, concluding comments are given in Section 5.

## II. PROBABILISTIC NEURAL NETWORKS IN IMAGE RECOGNITION

In image recognition task it is required to assign an observed image $X$ to one of $C > 1$ classes. The $c$-th class is specified by a set of $R(c) \geq 1$ reference images (instances) $X_r(c)$, $r \in \{1, ..., R(c)\}$. The total size of the training set is equal to $R = \sum_{c=1}^{C} R(c)$. We will assume, that this set is not very large, so it is not suitable for training a deep CNN from scratch. Hence, the variants of transfer learning are typically applied nowadays. At first, each image is described with a feature vector. The feature extraction is implemented with the deep CNN [1]. The CNN is pre-trained with an external large dataset, e.g., ImageNet [54] or CASIA-WebFace [55]. After that, the CNN can be fine-tuned to the given training set $\{X_r(c)\}$[56]. In fact, it is not necessary to use the neural network as a final classifier. Any appropriate machine learning technique can be applied to the features extracted by the original CNN. In this case, the outputs of one of the last (bottleneck) layers of this CNN for the input image $X$ and each $r$-th reference image are used as the feature vectors. After that, this vectors are $L_2$-normed in order to prevent the known variability to the observation conditions. As a result, $D \gg 1$ features $\mathbf{x}_r(c) = [x_{r;1}(c), ..., x_{r;D}(c)]$ are extracted from the $r$-th training image of the $c$-th class to obtain the instance. The outputs of the deep CNNs at the last layers are usually uncorrelated, though the PCA can be used further to reduce the dimensionality and make the features to be uncorrelated. The same procedure is applied to an observed image, and the $D$-dimensional feature vector $\mathbf{x} = [x_1, ..., x_D]$ is obtained.

Let us discuss the usage of statistical approach to perform the classification. This approach assumes that the feature vectors from one class have the same probability distribution, and the task is reduced to the testing of $C$ hypothesis $W_c, c = \overline{1, C}$ about distribution of the observation $\mathbf{x}$ [20]. If the prior probabilities of each class are computed based on the number of class instances in the training set $R(c)$, the maximum a posteriori (MAP) decision is written as follows:

$$c^* = \underset{c \in \{1, ..., C\}}{\arg\max} \frac{R(c)}{R} f(\mathbf{x}|W_c). \qquad (1)$$

Here $f(\mathbf{x}|W_c)$ is the conditional probability density function (likelihood) of the hypothesis $W_c$. This likelihood is estimated with the given training set by using, e.g., the non-parametric techniques [36]:

$$\hat{f}(\mathbf{x}|W_c) = \frac{1}{R_c} \sum_{r=1}^{R(c)} K(\mathbf{x}, \mathbf{x}_r(c)), \qquad (2)$$

where $K(\mathbf{x}, \mathbf{x}_r(c))$ is an appropriate kernel function. The classifier (1), (2) implements the well-known PNN [38], if the Gaussian Parzen kernel is applied

$$K(\mathbf{x}, \mathbf{x}_r(c)) = \frac{1}{(2\pi\sigma^2)^{D/2}} \exp\left(-\frac{1}{2\sigma^2} \sum_{d=1}^{D} (x_d - x_{r;d}(c))^2\right). \qquad (3)$$

Here $\sigma$ is the smoothing parameter. Though this classification algorithm (1)-(3) can be learned very fast just by memorizing the whole dataset, its run-time complexity is equal to O($DR$), and the space (memory) complexity also linearly depends on the size of the training set. This drawback obviously restricts the potential usage of the PNN. In the next section we describe the possibility to increase its performance by using other kernel functions based on the orthogonal series [48], [49].

## III. PROBABILISTIC NEURAL NETWORKS WITH ORTHOGONAL SERIES

### A. Fourier-series Estimates

It is known [35], that the densities in (2) can be estimated with the orthogonal series kernel (trigonometric, Hermite, Legendre, etc.) instead of the Gaussian-Parzen kernel (3). However, the computational complexity of such procedure for one-dimensional feature remains identical to the original PNN. Moreover, the complexity becomes exponential for high-dimensional feature vectors, if the joint distributions are estimated. However, in our task the outputs of the deep CNN are known to be approximately uncorrelated, so that it is possible to assume the independence of individual features. Hence, the naïve Bayesian rule can be used [20] to estimate the joint likelihood in (1) as the product of individual feature densities $\hat{f}_d(x_d|W_c)$:

$$\hat{f}(\mathbf{x}|W_c) = \prod_{d=1}^{D} \hat{f}_d(x_d|W_c), \qquad (4)$$

where the probability density function of the $d$-th feature is written similarly to (2):

$$\hat{f}_d(x_d|W_c) = \frac{1}{R(c)} \sum_{r=1}^{R(c)} K(x_d, x_{r;d}(c)). \qquad (5)$$

If the series of orthogonal functions $\{\psi_j(x)\}, j = 0, 1, 2, ...$ is used to estimate the unknown density (5), the cut-off parameter $J$ should be chosen to truncate the orthogonal series [49]. Hence, the kernel in (5) is approximated with the following expression:

$$K(x_d, x_{r;d}(c)) = \sum_{j=1}^{J} \psi_j(x_d) \psi_j(x_{r;d}(c)). \qquad (6)$$

As the feature vector was $L_2$-normalized, the support of every $d$-th feature is bounded: $x_d \in [-1, 1]$. In this case it is recommended to use the complex exponential orthogonal functions:

$$\psi_j(x) = \exp(ij\pi x), j = 0, \pm 1, \pm 2, ..., \qquad (7)$$

where $i = \sqrt{-1}$ is the imaginary unit. Hence, the Gaussians (3) are replaced by complex exponential activation functions (7).

Unfortunately, performance of this *canonical* form of the orthogonal series-based estimates (5) is $J$-times worse when compared with the PNN (1)-(3) due to the need for compute the kernel between each individual feature (6). However, it is easy to show, that the function (6), (7) is equivalent to the Dirichlet kernel

$$D_J(x_d, x_{r;d}(c)) = \frac{\sin\left(\left(J + \frac{1}{2}\right)\pi(x_d - x_{r;d}(c))\right)}{\sin\left(\frac{1}{2}\pi(x_d - x_{r;d}(c))\right)}. \qquad (8)$$

Hence, the individual density is estimated as follows:

$$\hat{f}_{J;d}(x_d|W_c) = \frac{1}{R(c)} \sum_{r=1}^{R(c)} D_J(x_d, x_{r;d}(c)). \qquad (9)$$

Here we added index $J$ into $\hat{f}_{J;d}(x_d|W_c)$ to emphasize its dependence on the cut-off. Thus, the complexity of the classifier (1), (4), (9) based on the orthogonal series remains linear – O($DRJ$) – and is asymptotically equivalent to the original PNN [38].

### B. Proposed Approach

At first, it is necessary to highlight the limitation of the trigonometric series (7) applied to the estimation of the probability density function. Namely, the Dirichlet kernel (8) is not always non-negative. To prevent this drawback, in this paper we estimate the individual likelihood in (4) as the *average* of the first $J$ partial sums (9):

$$\hat{f}_d\left(x_d|W_c\right) = \frac{1}{J+1}\sum_{j=0}^{J}\hat{f}_{j;d}\left(x_d|W_c\right). \quad (10)$$

It is easy to show that this expression can be written in the canonical form (5) with the (always non-negative) Fejér kernel:

$$F_{J+1}\left(x_d, x_{r;d}(c)\right) = \frac{1}{J+1}\frac{1-\cos\left((J+1)\pi\left(x_d - x_{r;d}(c)\right)\right)}{1-\cos\left(\pi\left(x_d - x_{r;d}(c)\right)\right)}. \quad (11)$$

This kernel is widely used in the orthogonal series expansions [48]. To simplify the computations further, we replace the equation (9) to the equivalent non-canonical form [35]:

$$\hat{f}_{J;d}\left(x_d|W_c\right) = \sum_{j=-J}^{J}a_{j;d}(c)\exp\left(ij\pi x_d\right), \quad (12)$$

where the Fourier coefficients $\left\{a_{j;d}(c)\right\} j\in\{-J,...,J\}, d\in\{0,1,...,D\}$ are computed using the given training set as follows:

$$a_{j;d}(c) = \frac{1}{R(c)}\sum_{r=1}^{R(c)}\exp\left(ij\pi x_{r;d}(c)\right). \quad (13)$$

By substituting equation (12) to the mean estimate (10), we obtain the final expression:

$$\hat{f}_d\left(x_d|W_c\right) = \sum_{j=-J}^{J}\frac{J+1-|j|}{J+1}a_{j;d}(c)\exp\left(ij\pi x_d\right). \quad (14)$$

This expression makes it possible to reduce the classification performance to O($JD$). This is usually a dramatic reduction, because the cut-off $J$ is much lower when compared with the size of the training set. For example, it is known [47], that the estimation (14) converges to the real probability distribution with the convergence rate $O\left((R_c)^{-2/3}\right)$, if the number of terms (cut-off) $J$ is equal to $O\left((R_c)^{1/3}\right)$ [50], [57]. In practice, this parameter is chosen experimentally by minimizing the mean integrated square error [49]. For instance, Hart proposed the data-driven procedure to define cut-off $J_d^*(c)$[58], which maximizes the following expression

$$\sum_{j=1}^{J}\left(a_{j;d}(c)\right)^2 - \frac{2J}{R_c+1}. \quad (15)$$

Unfortunately, our experiments show that if the cut-offs $J_d^*(c)$ depend on the class label, the recognition accuracy will be rather low. We decided that it is necessary to use the same cut-off in order to properly compare the estimated likelihoods in (1). Thus, we used the following heuristic. After computation of the optimal cut-offs $J_d^*(c)$ (15) we calculate their median $J*$ and used it as the value of parameter $J$ in estimation of likelihoods (14) for *all* classes and features.

### C. Implementation Details

Let us describe the proposed recognition procedure with several further details. During the training procedure, the neural network weights $\left\{w_{j;d}(c)\right\} j\in\{0,1,...,J*\}, d\in\{0,1,...,D\}$ are computed using

the Fourier coefficients (13):

$$w_{j;d}(c) = \frac{J+1-j}{J+1}a_{j;d}(c) =$$
$$= \frac{J+1-j}{(J+1)R(c)}\sum_{r=1}^{R(c)}\exp\left(ij\pi x_{r;d}(c)\right) \quad (16)$$

Next, the basis functions (7) are computed for each feature of the observed image. Though this procedure is quite computationally expensive, it can be simplified by using the known trigonometric rules:

$$\psi_{j+1}(x_d) =$$
$$= \mathrm{Re}\{\psi_{j-1}(x_d)\}\cdot\mathrm{Re}\{\psi_1(x_d)\} - \mathrm{Im}\{\psi_{j-1}(x_d)\}\cdot\mathrm{Im}\{\psi_1(x_d)\} + \quad (17)$$
$$+ i\Big(\mathrm{Re}\{\psi_{j-1}(x_d)\}\cdot\mathrm{Im}\{\psi_1(x_d)\} + \mathrm{Im}\{\psi_{j-1}(x_d)\}\cdot\mathrm{Re}\{\psi_1(x_d)\}\Big)$$

These recursive computations are initialized as follows

$$\psi_1(x_d) = \cos\left(\pi x_d\right) + i\sin\left(\pi x_d\right) \quad (18)$$

Hence, the expensive trigonometric functions are computed only twice in the recursion initialization (18) for each feature of observed image. After that, the individual likelihoods are estimated (14). As the resulted probabilities can be rather low, we convert them to the log-scale and maximize the log-likelihood:

$$c* = \operatorname*{argmax}_{c\in\{1,...,C\}}\sum_{d=1}^{D}\log\sum_{j=-J}^{J}w_{j;d}(c)\cdot\psi_j(x_d) + \log R_c, \quad (19)$$

The proposed approach (16)-(19) saves all advantages of the PNN [38]: it converges to the optimal Bayesian decision (1) and can be implemented completely in parallel. The new instance can be added even in real time. The following equation is used to refine the weights (16) of the network, if the new instance $\mathbf{x}_{R(c)+1}(c)$ of the $c$-th class is available:

$$\widetilde{w}_{j;d}(c) = \frac{R(c)}{R(c)+1}w_{j;d}(c) +$$
$$+ \frac{J+1-j}{(J+1)(R(c)+1)}\exp\left(ij\pi\cdot x_{R(c)+1;d}(c)\right). \quad (20)$$

The run-time complexity of such modification of the training set is equal to $O(D\cdot J*)$. It is slightly more expensive, than just an addition of a node to the pattern layer in the original PNN, but it is appropriate in practical applications. What is more important, the proposed approach provides much more faster on-line classification performance. We provide the evidence in the next subsection.

### D. Proposed Algorithm

The complete recognition algorithm (16)-(20) is presented in Table I. Here we store the real and imaginary parts of the complex weights (16) and preprocessed input features (17), (18) in independent arrays using trigonometric functions. The orthogonal series estimate [50], [57] should theoretically converge if the cut-off parameter $J$ is proportional to the cubic root of the number of samples. Hence, the runtime complexity of the density estimates (12) with the trigonometric series and a cut-off value can be defined as $O\left((R(c))^{1/3}\right)$. As a result, the most important advantage of our algorithm (Table I) over the PNN (1)-(3) is approximately the $(R/C)^{2/3}$–times improved performance of online pattern recognition, if each class contains the same number of reference instances. Moreover, the space complexity also decreases to $O\left(CD\cdot(R/C)^{1/3}\right)=O\left(D\cdot R^{1/3}C^{2/3}\right)$, because only $C\cdot D\cdot(2J*+1)$ weights (16) are stored. Thus, such a modification of the PNN does not implement the memory-based classifier anymore [27].

TABLE I
PROPOSED IMAGE RECOGNITION ALGORITHM

**Data**: observed image, training instances with given class labels
**Parameters**: cut-off $J$; deep CNN, which has already been trained to classify images from very-large dataset
**Output**: class label $c^*$ (19)
**Preliminary steps**:
1. For each class label $c$=1,2,…,$C$ repeat
1.1. For each instance $r$=1,2,…,$R(c)$ repeat
1.1.1. Feed the $r$-th instance into the input layer of the CNN and compute $D$ outputs of a bottleneck layer
1.1.2. Compute $L_2$ norm of these outputs and normalize them to obtain the feature vector $\mathbf{x}_r(c)$
1.2. For each dimension $d$=1, 2, …, $D$ repeat
1.2.1. For each weight $j$=1,2, … , $J$ repeat // compute Eq. (16)
1.2.1.1. Assign $w_{\cos}[c,j,d]:=0; w_{\sin}[c,j,d]:=0$
1.2.1.2. For each instance $r$=1,2,…,$R(c)$ repeat
1.2.1.2.1. $w_{\cos}[c,j,d]:=w_{\cos}[c,j,d]+\dfrac{J+1-j}{(J+1)R(c)}\cos\left(j\pi x_{r;d}(c)\right)$
1.2.1.2.2. $w_{\sin}[c,j,d]:=w_{\sin}[c,j,d]+\dfrac{J+1-j}{(J+1)R(c)}\sin\left(j\pi x_{r;d}(c)\right)$
1.2.2. Assign $w_{\cos}[c,0,d]:=1/R(c)$

**Online recognition steps**:
1. Feed the observed image into the input layer of the CNN and compute $D$ outputs of a bottleneck layer
2. Compute $L_2$ norm of these outputs and normalize them to obtain the feature vector $\mathbf{x}$
3. Initialize two $J*$-dimensional arrays $\psi_{\cos}[], \psi_{\sin}[]$
4. For each class $c$=1, 2, …, $C$ repeat // Initialize output array
5. Assign $Log[c]:=\log R(c)$
6. For each dimension $d$=1, 2, …, $D$ repeat // compute Eq. (17), (18)
6.1. Assign $\psi_{\cos}[1]:=\cos(\pi x_d), \psi_{\sin}[1]:=\sin(\pi x_d)$
6.2. For each $j$=2,3,…,$J$ repeat
6.2.1. Assign $\psi_{\cos}[j]:=\psi_{\cos}[j-1]\psi_{\cos}[1]-\psi_{\sin}[j-1]\psi_{\sin}[1]$
6.2.2. Assign $\psi_{\sin}[j]:=\psi_{\cos}[j-1]\psi_{\sin}[1]+\psi_{\sin}[j-1]\psi_{\cos}[1]$
6.3. Assign $Out:=w_{\cos}[c,0,d]/2$
6.4. For each $j$=1,2, …, $J$ repeat // compute Eq. (19)
6.4.1. $Out:=Out+w_{\cos}[c,j,d]\cdot\psi_{\cos}[j]+w_{\sin}[c,j,d]\cdot\psi_{\sin}[j]$
6.5. Assign $Log[c]:=Log[c]+\log(Out)$
7. Return the index $c^*$ of the maximum value of the output array $Log[]$.

The proposed PNN with the complex exponential activations functions is presented in Fig.1. In this modification we implemented $J*$ new kernel layers to compute the complex exponential activation functions (17), (18). The next, pattern, layer does not include anymore the comparison of new observation and *all* instances, as it is done in the original PNN with the Guassian Parzen kernel (3). On the contrary, only a sum of $C\cdot D\cdot(2J*+1)$ products with the weights (16) are computed in this layer in Fig. 1. Thus, the number of neurons at the pattern layer of our network is much less when compared to $R$ neurons in the classical PNN. This dramatic reduction is achieved by using the outputs of $2DJ*$ neurons with trigonometric (or complex exponential) activations functions (17), (18) instead of the matching of original feature vectors in the Gaussian kernel (3) of the PNN. The log-likelihood for each class is estimated in the summation layer. The last output (or decision) layer produces the class label with the maximal posterior probability (19).

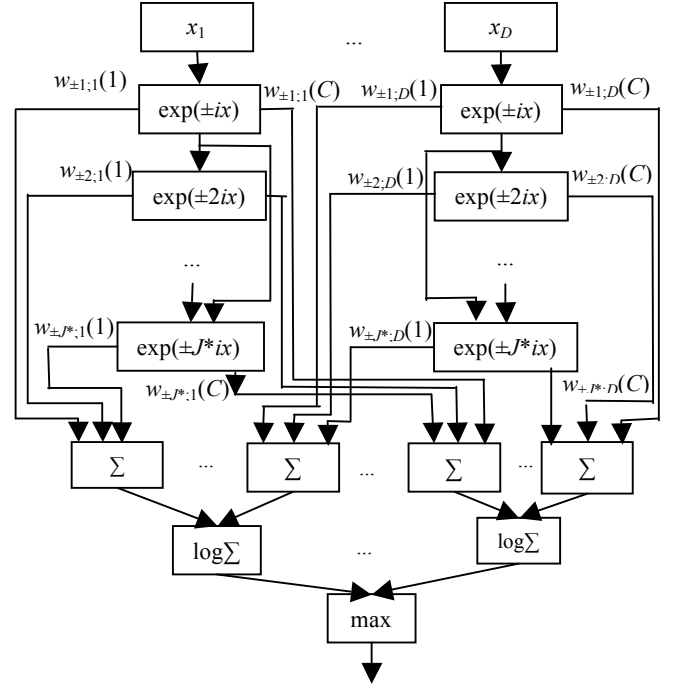

Fig. 1. Probabilistic neural network with the complex exponential activations functions

In fact, though our approach is based on the implementation of the PNN with the orthogonal series kernel functions [35], our network (Fig. 1) does not have much in common with the original PNN. At first, we use additional $J*$ simple layers with $2D$ neurons in each layer in order to transform the feature vectors to the trigonometric space (7). This procedure simplifies the final criterion (19) based on the non-canonical form of the densities estimates (12). The latter includes only the linear combination of the outputs of the previous layers. Secondly, the PNN stores all features of all instances in the weights of the pattern layer. On the contrary, our network implements some kind of distributed representation [1], so that each weight (16) in the pattern layer contains information about all instances of particular class (20).

To draw the line, the usage of the complex exponential activation functions (7) instead of traditional Gaussians (3) made it possible to estimate the probability densities using the Fejér kernel (11) using non-canonical form (12), which does not require the brute force (5) of all instances from the training dataset. As a result, our network is expected to improve performance and space complexity of the PNN in approximately $(R/C)^{2/3}$ times. The next section experimentally supports the benefits of the proposed approach.

## IV. Experimental Results

In this section we compare the proposed modified network (Fig. 1) with the original PNN (1)-(3) and the known reduction of the PNN [44], which chooses several centroids in each class with the k-median clustering method and estimates the densities (2) using only these centroids [20]. In addition, we examine the k-nearest neighbor (k-NN) method, the nearest centroid (or Rocchio) method [59], which is very similar to the PNN's reduction [44] and other multi-class classifiers implemented in the OpenCV library, namely, random forest (RF), MLP with one hidden layer, and SVM with linear and RBF (Radial Basis Functions) kernels [20], [22].

The deep learning Caffe framework [60] is applied to extract image features. Performance of these methods is estimated on a laptop (4 core 2 GHz Intel Core i7, 6Gb RAM) by using a console application (https://github.com/HSE-asavchenko/HSE_FaceRec/tree/master/src/recognition_testing) built in Qt 5 environment with Microsoft Visual C++ 2013 compiler (x64 environment) and optimization by speed.

The testing protocols for the datasets used in our experiments were specially designed to compare the accuracy of recognition methods for rather large training samples. That is why we implemented the following variant of the random subsampling cross-validation, which is suitable to examine the small-sample-size case. Every dataset is divided randomly 10 times into the training and testing sets, so that the ratio of the size $R$ of the training set to the size of the whole dataset $T$ is equal to a fixed number. We split each class independently in order to provide approximately identical ratio of the number of images in the training and testing set. In this paper we focus on the small sample size problem, so the ratio $R/T$ was chosen to be rather low. As all the datasets are imbalanced, we estimate the accuracy of recognizing images from each class individually and computed the mean accuracy (i.e., a recall, as the classifier always returns one class). In addition, the average time to classify one image is measured.

The following classifier parameters were tuned for each number of instances in the dataset: the smoothing parameter $\sigma$ (0.001, 0.005, 0.01, 0.015, …, 1.0) of the PNN, the number of centroids (1, 3, 5, 10) in the reduced PNN and the nearest centroid method, the size of the neighborhood k (1, 3, 5) in the k-NN, the number of hidden neurons in the MLP (64, 128, 256 and 512), the parameter C (0, 0.001, 0.01, 0.1, 1) in SVM. The cut-off parameter $J^*$ of our algorithm (Table I) was simply fixed to be equal to $2\max\left((R/C)^{1/3}, 1\right)$ in order to obtain an accurate estimation of the probability density for a very small training sample. Unfortunately, we cannot tune these parameters using only training sets because they are too small. Thus, we decided to use the extra datasets (Caltech-UCSD Birds 200 and the Labeled Faces in the Wild [61]) for parameters tuning in image categorization and face recognition tasks, respectively. Though it is possible to obtain better results after parameters tuning with the same dataset as the training/testing set, we believe, that it is not a frank procedure, because it introduces a bias into the testing results. At the same time, the absence of such tuning for each individual dataset can lead to slightly worse results, which are reported in literature.

### A. Object Category Recognition

The first group of experiments is devoted to the visual object category recognition task. We used three datasets with rather large number of classes:

1. Caltech 101 Object Category dataset [62], which contains $T = 8677$ images of $C$=101 classes, i.e., we ignored the distractor background class.

2. Caltech 256 dataset [63] with $T = 29780$ images of $C$=256 classes (without clutter class).

3. Stanford Dogs dataset [64] with $T = 20580$ images of $C = 120$ classes.

To extract image features in preliminary steps of our algorithm (Table I), traditional transfer learning techniques were applied. We downloaded two deep CNN architectures (19-layer VGGNet [7] and GoogLeNet [8]) from the Caffe Model Zoo repository. These CNNs have already been trained to recognize images of 1000 classes from ImageNet dataset [54]. After that each RGB image from each dataset (Caltech-101, Caltech-256 and Stanford Dogs) was fed into the input of these CNNs. The outputs of one of the last layers (pool5/7x7-s1 for GoogLeNet and fc6 for VGGNet) were $L_2$-normalized in order to produce the final feature vector with dimensionality $D$=1024 and $D$=4096 for GoogLeNet and VGGNet.

The dependence of the accuracy and classification time (in the format mean ± standard deviation) on the average number of instances per class for the Caltech-101 dataset and the GoogLeNet features is presented in Table II and Table III, respectively. The best results are marked by bold.

Here, firstly, conventional SVM is the most accurate classifier only when the training set is rather large (more than 20 images per class). However, it is one of the worst methods (together with RF and MLP) for very small sample size. Performance of the linear SVM is very low and practically does not depend on the training set size $R$, because this model cannot satisfy the termination criterion, so the maximum number of support vectors (5050 in OpenCV) was always chosen. In contrast to the linear case, SVM with the RBF kernel was able to converge, so that the number of support vectors becomes an order of magnitude lower. Hence, it is several times faster, than the linear SVM, even if the computation of RBF kernel is time-consuming. However, the accuracy on the testing set (Table II) of the linear kernel is slightly higher when compared to the RBF. This result is not trivial, but it is typical for practical classification tasks in high-dimensional feature space [59]. Secondly, the RF is the least accurate classifier in all cases, so we do not consider it in the most part of the next experiments. It is worth noting that RF and MLP are much faster than SVMs. The latter is a binary classifier, and the one-vs-all strategy is used by the OpenCV implementation in our multi-class task. That is why this matrix multiplication in linear SVMs should be repeated for each class. Thirdly, accuracies of the instance-based classifiers (k-NN and PNN) are approximately identical. As it is expected, their performance increases linearly when $R$ is increased. The reduction of the PNN with clustering and the nearest centroids improve classification speed of the baseline PNN and k-NN, if the number of clusters is much less than the size $R$ of the whole training set. However, the clustering causes an increase of accuracy on 2-3%, because the resulted decision looses the property of the Bayesian optimality. Finally, the proposed modification of the PNN (Fig. 1) is in most cases the best classifier. It is characterized with the highest accuracy for the small samples. For instance, it is 3-4% more accurate, than the PNN and k-NN, and 25-35% more accurate, than the state-of-the-art classifiers (RF, MLP and SVM), if the training set contains only 5 images for each class. Our modification is also one of the fastest classifiers. In fact, it is more preferable, than the original PNN, in all cases: our accuracy is 2-4% higher, and the classification speed is 1.5-3.8-times lower. It is important to note, that the classification time varies slightly with the linear increase of the size $R$ of the training set.

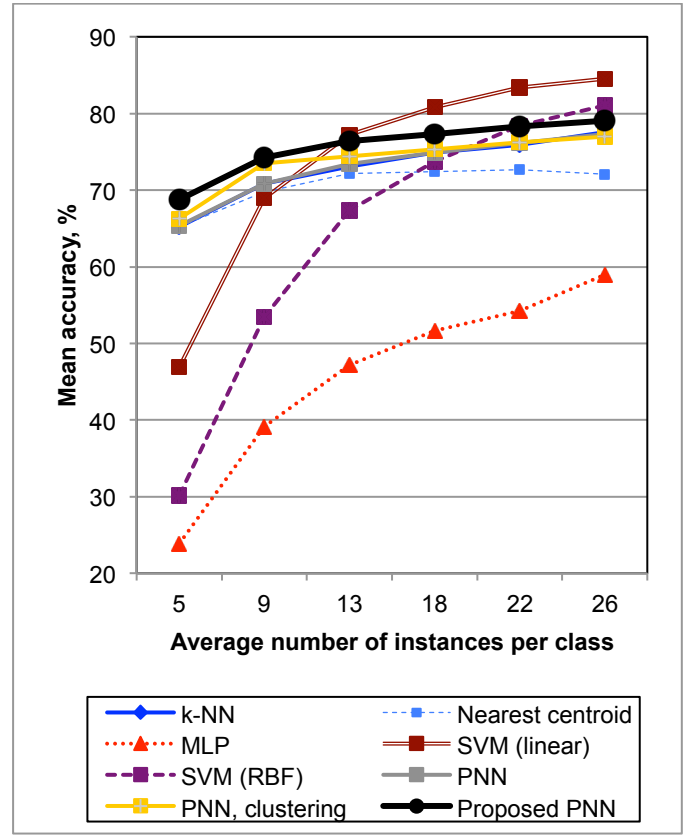

Fig. 2. Mean accuracy (%), Caltech101 dataset, VGGNet features

TABLE II
MEAN ACCURACY (%), CALTECH-101, GOOGLENET FEATURES

| Classifier | Average number of instances per class | | | |
|---|---|---|---|---|
| | 5 | 13 | 22 | 26 |
| k-NN | 70.1±0.3 | 76.8±0.2 | 78.7±0.3 | 79.8±0.4 |
| Nearest centroid | 70.1±0.2 | 75.9±0.2 | 76.0±0.3 | 76.9±0.1 |
| RF | 38.2±0.7 | 67.8±0.3 | 73.0±0.6 | 74.4±0.6 |
| MLP | 39.5±6.8 | 73.0±0.3 | 79.1±1.7 | 81.3±3.5 |
| SVM (linear) | 49.9±1.1 | 78.3±0.4 | **82.8±0.3** | **83.9±0.2** |
| SVM (RBF) | 42.4±0.3 | 75.4±0.7 | 81.4±0.3 | 83.8±0.5 |
| PNN | 70.5±0.5 | 77.6±0.2 | 79.4±0.2 | 80.9±0.6 |
| PNN, clustering | 71.3±0.5 | 76.9±0.9 | 77.2±0.5 | 77.8±0.8 |
| **Proposed PNN** | **74.2±0.5** | **80.2±0.4** | 81.9±0.4 | 82.3±0.6 |

TABLE III
RECOGNITION TIME (MS), CALTECH-101, GOOGLENET FEATURES

| Classifier | Average number of instances per class | | | |
|---|---|---|---|---|
| | 5 | 13 | 22 | 26 |
| k-NN | 1.0±0.1 | 2.8±0.1 | 4.7±0.2 | 5.7±0.3 |
| Nearest centroid | 0.8±0.1 | 1.1±0.1 | 1.1±0.1 | 1.1±0.1 |
| RF | **0.2±0.1** | 0.3±0.1 | 0.4±0.2 | 0.4±0.2 |
| MLP | **0.2±0.1** | **0.2±0.1** | **0.2±0.1** | **0.2±0.1** |
| SVM (linear) | 3.7±0.2 | 3.8±0.2 | 3.8±0.2 | 3.8±0.3 |
| SVM (RBF) | 0.6±0.1 | 1.7±0.2 | 2.5±0.3 | 3.0±0.3 |
| PNN | 1.0±0.1 | 2.9±0.1 | 4.6±0.2 | 5.5±0.4 |
| PNN, clustering | 0.7±0.1 | 1.0±0.1 | 1.1±0.1 | 1.1±0.2 |
| **Proposed PNN** | 1.0±0.1 | 1.4±0.2 | 1.5±0.3 | 1.5±0.3 |

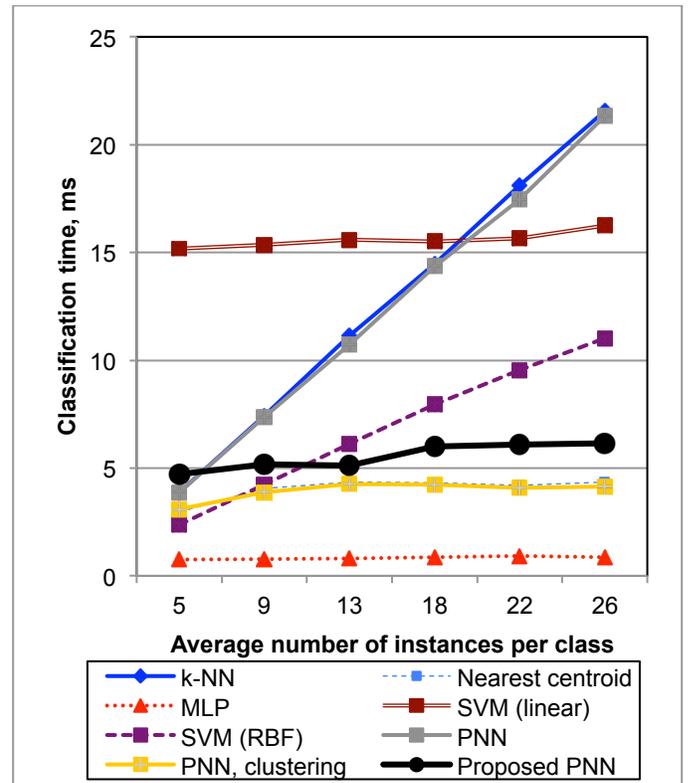

Fig. 3. Classification time (ms), Caltech101 dataset, VGGNet features

In the next experiment we used the same dataset, but with features extracted by the VGGNet. The results (Fig. 2, 3) are very similar to the previous experiments, but the classification complexity is higher, and the gain in performance of the proposed approach (Table I) is more noticeable.

Here our modification is 2-3% more accurate and 1.4-3.4 times faster, when compared with the original PNN, if the training set contains no less than 9 images per class. The classification time of the linear SVM is 2.5-3.2-times higher than the recognition time for our network (Fig. 1). However, SVM with the linear kernel is more accurate even for rather small training samples (13 instances per class). It is remarkable, that the accuracy of SVM for these features is 0.8% higher than its accuracy for the GoogLeNet features (Table II). At the same time, the accuracy of all other methods in this experiment decreases at 3-6% when compared to the GoogLeNet features.

In the next experiment much more complex Caltech 256 dataset is explored. The mean accuracy and classification time for features from the GoogLeNet are shown in Fig. 4, 5. Conventional linear SVM is the most accurate, but very slow classifier for rather large samples. Besides MLP, which performance does not depend on the database size, the most efficient methods (PNN with clustering and the nearest centroid) include preliminary clustering of the training set. At the same time, their accuracies are up to 7% lower when compared to the original PNN and baseline k-NN. Finally, the proposed modification of the PNN (Fig. 1) is again one of the most efficient classifiers. For example, its accuracy is 3-8% higher, than the accuracy of the PNN (1)-(3) and the k-NN. Its speed is also rather high: an observed image is recognized 1.5-3.3-times faster, than with the PNN. What is important, performance of our classifier (15)-(20) does not decrease linearly with the increase of the size of the training set $R$.

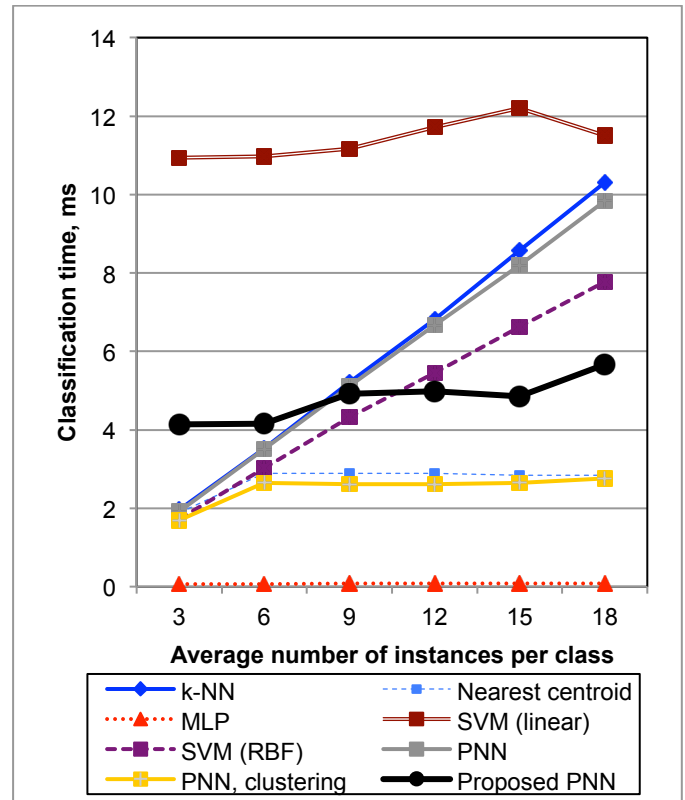

Fig. 5. Classification time (ms), Caltech256 dataset, GoogLeNet features

In the next experiment we present several results obtained for the Stanford Dogs dataset (Fig. 6, 7).

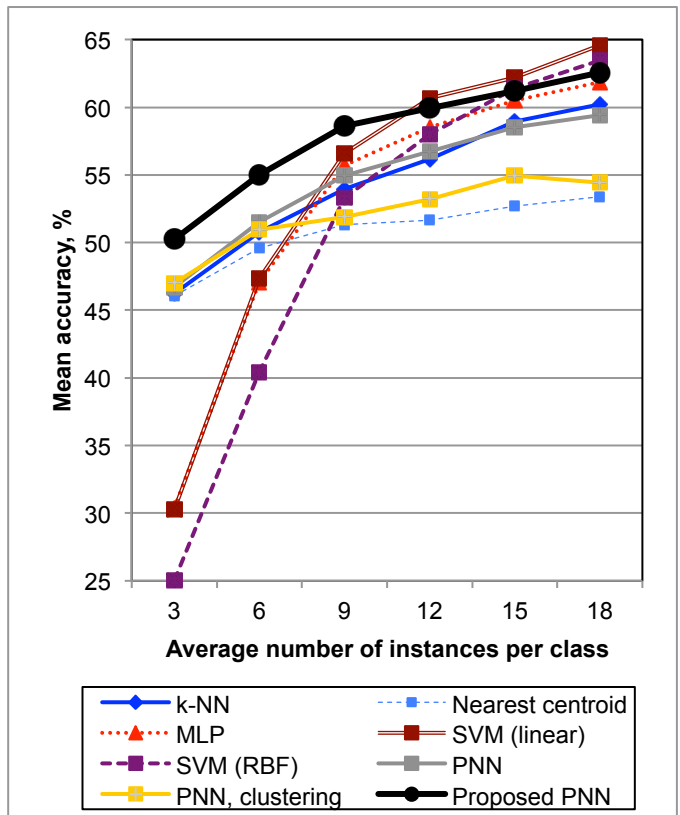

Fig. 4. Mean accuracy (%), Caltech256 dataset, GoogLeNet features

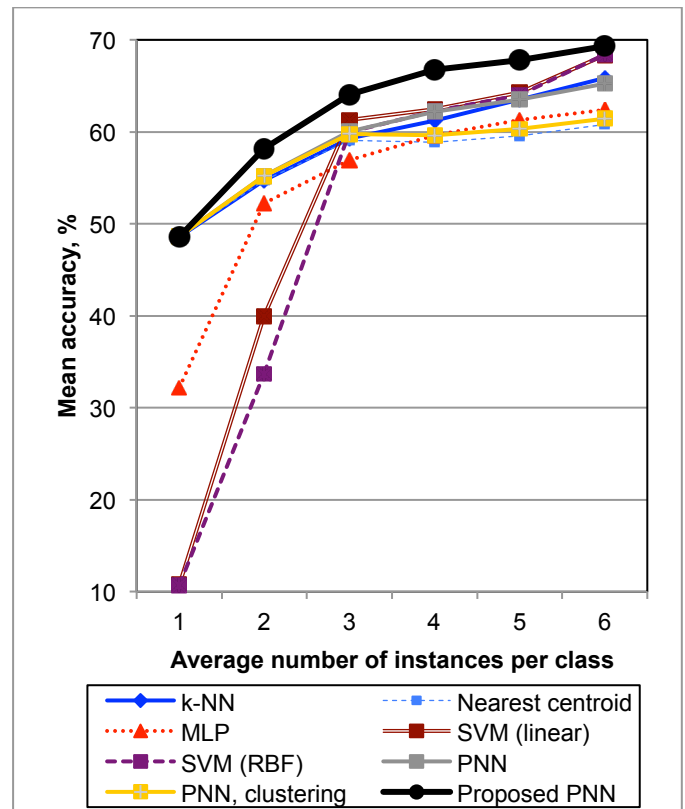

Fig. 6. Mean accuracy (%), Stanford Dogs dataset, GoogLeNet features

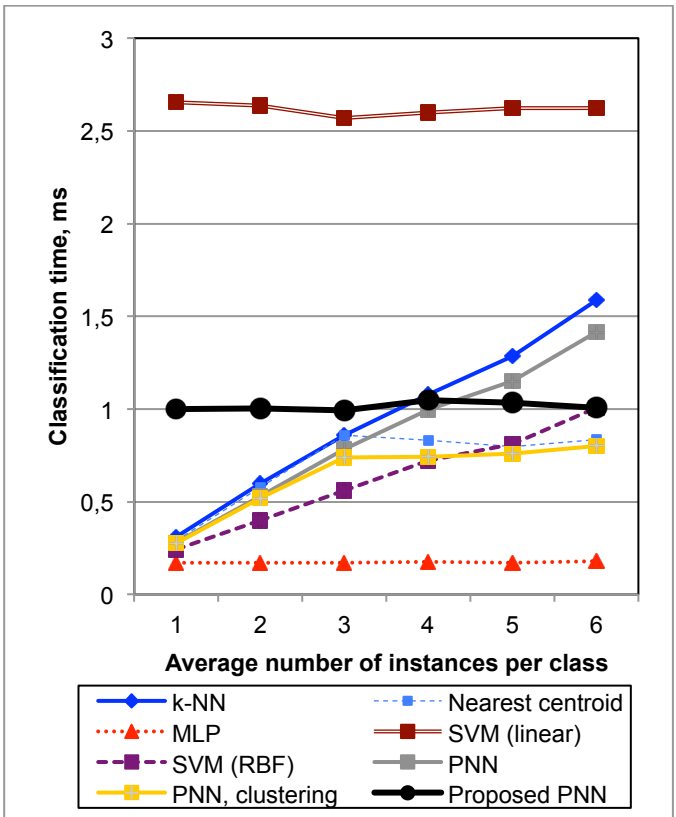

Fig. 7. Classification time (ms), Stanford Dogs dataset, GoogLeNet features

| Dataset | Classifier | GoogLeNet | | VGGNet | |
|---|---|---|---|---|---|
| | | No PCA ($D$=1024) | PCA ($D$=128) | No PCA ($D$=4096) | PCA ($D$=256) |
| Caltech-101 | k-NN | 74.1 | 74.5 | 70.8 | 71.0 |
| | Nearest centroid | 73.8 | 73.7 | 69.8 | 69.6 |
| | RF | 51.9 | 62.3 | 36.1 | 50.1 |
| | MLP | 60.0 | 75.0 | 39.1 | **75.7** |
| | SVM (linear) | 70.3 | 64.5 | 69.0 | 63.9 |
| | SVM (RBF) | 65.5 | 68.9 | 53.4 | 65.0 |
| | PNN | 74.6 | 74.0 | 70.8 | 69.9 |
| | PNN, clustering | 74.9 | 73.4 | 73.5 | 69.7 |
| | **Proposed PNN** | **78.1** | **78.7** | **74.2** | 74.8 |
| Caltech-256 | k-NN | 53.5 | 53.3 | 52.8 | 53.1 |
| | Nearest centroid | 51.3 | 50.5 | 52.4 | 52.1 |
| | RF | 23.8 | 36.5 | 15.9 | 29.6 |
| | MLP | 55.7 | 52.2 | 38.5 | 54.7 |
| | SVM (linear) | 56.6 | 50.4 | 56.2 | 50.4 |
| | SVM (RBF) | 53.3 | 54.3 | 47.1 | 52.7 |
| | PNN | 54.9 | 53.2 | 52.9 | 53.0 |
| | PNN, clustering | 51.9 | 50.3 | 51.4 | 49.9 |
| | **Proposed PNN** | **58.6** | **59.4** | **56.8** | **57.3** |
| Stanford Dogs | k-NN | 66.5 | 66.6 | 65.8 | 66.2 |
| | Nearest centroid | 63.4 | 61.6 | 61.8 | 61.7 |
| | RF | 72.4 | 70.7 | 72.8 | 70.6 |
| | MLP | 63.4 | 64.8 | 40.5 | 66.7 |
| | SVM (linear) | 74.8 | 74.3 | **74.7** | **74.2** |
| | SVM (RBF) | **75.1** | **74.8** | 74.4 | 74.4 |
| | PNN | 68.4 | 66.7 | 66.0 | 66.3 |
| | PNN, clustering | 64.0 | 61.1 | 61.9 | 61.8 |
| | **Proposed PNN** | 72.9 | 73.4 | 72.2 | 72.6 |

This task is the most challenging when the training sample is extremely small, so here we consider only 1, 2, ..., 6 instances for each class. That is why the proposed algorithm (Table II) is the most accurate one in all cases (Fig. 6). Its error rate is 3-4% lower, when compared to the PNN. Our algorithm is up to 9% more accurate than the PNN with clustering. However, it is slower than the PNN if only 4 images are available for each dog type. However, even in this case our approach takes only 1 ms to recognize one image, which is appropriate in most practical case. Moreover, in contrast to most of the other methods, classification time for the proposed modification (Fig. 1) practically does not depend on the training set size.

In order to make all object categorization results of this subsection more clear, we decided to summarize the accuracy and recognition time in Tables IV and Table V, respectively. We consider only one, rather typical case, in which the training set contains 10 images for each class. Here we examined both known ways to make the PNN faster, namely, the selection of the most representative prototypes ("PNN, clustering") and the dimensionality reduction with the PCA. In particular, we compute 128 and 256 principal components of features extracted by the GoogLeNet and VGGNet, respectively.

Here one can notice that the VGGNet features lead to less accurate results when compared to the GoogLeNet features. Secondly, application of PCA causes much faster recognition. Moreover, the accuracy also slightly increases in most cases, though the error rate after application of PCA mostly decreases in our experiments with the lower training set. Thirdly, the problem of the small samples cannot be simply defined by the database size. For example, 10 images per class allow traditional classifiers (SVM, RF) to achieve relatively high accuracy for the Stanford Dogs dataset, but such amount of reference instances is absolutely not enough for these methods to recognize images from complex Caltech-256 dataset. Finally, these results are presented to demonstrate that our algorithm (Table I) is not the best classifier in all cases (see, e.g., Stanford Dogs dataset). However, we claim that the proposed modification (Fig. 1) allows to significantly decrease the image recognition time of the original PNN. In contrast to its known modifications, our approach is additionally characterized by the lower error rate. To conclude, the proposed algorithm is the most suitable method if the number of instances for each class is not large enough to train the traditional classifiers. For example, our modification of the PNN is the most accurate method even for the Stanford Dogs dataset, if there are at most 5 images of each class in the training set (Fig. 6).

TABLE V
IMAGE RECOGNITION TIME (MS), 10 INSTANCES PER CLASS

| Dataset | Classifier | GoogLeNet | | VGGNet | |
|---|---|---|---|---|---|
| | | No PCA ($D$=1024) | PCA ($D$=128) | No PCA ($D$=4096) | PCA ($D$=256) |
| Caltech-101 | k-NN | 1.94 | 0.33 | 7.40 | 0.59 |
| | Nearest centroid | 1.08 | 0.18 | 4.06 | 0.31 |
| | RF | 0.37 | 0.32 | **0.29** | 0.42 |
| | MLP | **0.07** | **0.05** | 0.79 | **0.07** |
| | SVM (linear) | 3.89 | 0.56 | 15.36 | 1.02 |
| | SVM (RBF) | 1.17 | 0.31 | 4.22 | 0.43 |
| | PNN | 1.89 | 0.28 | 7.37 | 0.53 |
| | PNN, clustering | 1.01 | 0.14 | 3.88 | 0.28 |
| | **Proposed PNN** | 1.31 | 0.16 | 5.18 | 0.30 |
| Caltech-256 | k-NN | 5.20 | 1.11 | 19.91 | 2.04 |
| | Nearest centroid | 2.89 | 0.54 | 10.75 | 1.08 |
| | RF | 0.56 | 0.56 | **0.68** | 0.62 |
| | MLP | **0.08** | **0.07** | 0.82 | **0.08** |
| | SVM (linear) | 11.17 | 3.47 | 34.2 | 6.99 |
| | SVM (RBF) | 4.32 | 1.75 | 15.60 | 2.81 |
| | PNN | 5.12 | 0.90 | 20.14 | 2.01 |
| | PNN, clustering | 2.62 | 0.45 | 11.23 | 0.97 |
| | **Proposed PNN** | 4.82 | 0.65 | 15.95 | 1.20 |
| Stanford Dogs | k-NN | 2.62 | 0.49 | 9.79 | 0.83 |
| | Nearest centroid | 0.77 | 0.13 | 3.06 | 0.23 |
| | RF | 0.29 | 0.31 | **0.31** | 0.44 |
| | MLP | **0.17** | **0.05** | 0.64 | **0.07** |
| | SVM (linear) | 5.18 | 0.72 | 11.93 | 3.16 |
| | SVM (RBF) | 1.74 | 0.45 | 5.85 | 0.65 |
| | PNN | 2.55 | 0.39 | 9.75 | 0.76 |
| | PNN, clustering | 0.74 | 0.11 | 2.92 | 0.20 |
| | **Proposed PNN** | 1.03 | 0.12 | 4.17 | 0.26 |

In the last experiment we demonstrate the tuning of the cut-off parameter $J$ (Table I). The dependence of the mean accuracy and classification time on the cut-off for the GoogLeNet features and 25 images per each class are shown in Fig. 8 and Fig. 9, respectively.

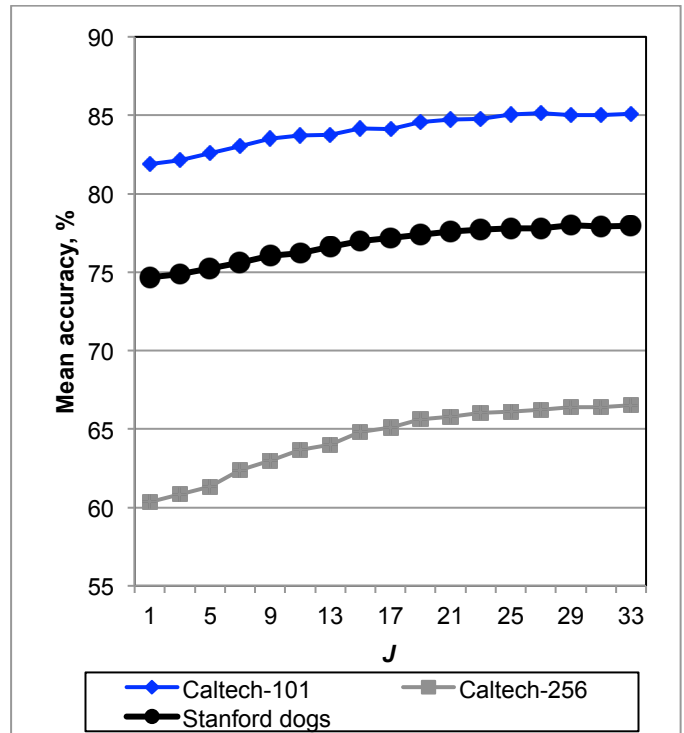

Fig. 8. Dependence of the mean accuracy (%) on the cut-off parameter $J$, GoogLeNet features, 25 images per each class

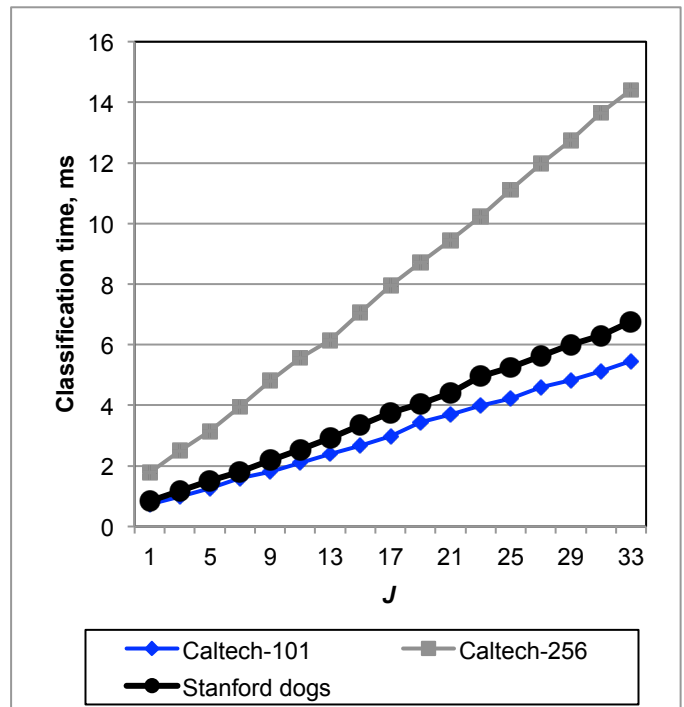

Fig. 9. Dependence of the classification time (ms) on the cut-off parameter $J$, GoogLeNet features, 25 images per each class

As it was expected, recognition time increases linearly (Fig. 9) with the increase of the cut-off $J$. The different number of classes in these datasets causes the different recognition time for the same $J$. The more is the cut-off, the better is the density estimate (12), and, hence, the higher is the recognition accuracy. However, the difference in the error rates becomes not statistically significant for each dataset, when $J$ exceeds a certain threshold (Fig. 8). We should note that in previous experiments the parameter $J$ was chosen to be much less than this threshold (e.g., $\left\lceil 2\max\left((R/C)^{1/3},1\right)\right\rceil = 6$ for 25 images per class). Such a cut-off provides an excellent performance and appropriate accuracy. Finally, the accuracy for Stanford Dogs recognition exceeds the baseline results achieved for matching of SIFT features [64] in 40-50%. To our knowledge, we obtained the state-of-the-art result for this dataset (Fig. 6), when its standard testing protocol with very the small training set (1-5 images per class) is used. As a matter of fact, such settings ideally fit the goals of our network (Fig. 1). The mean accuracy (recall) of our method for Caltech-101 (Fig. 8) is approximately identical to the results of the state-of-the-art standalone classifiers known from the literature [7], [65]. Thus, this experiment clearly demonstrates an advantage of our network (Fig. 1) over the baseline PNN: it is possible to find a compromise between performance and error rate by properly choosing the parameter $J$.

### B. Unconstrained Face Recognition

Another group of experiments is focused on unconstrained face identification [66]. We used the Lightened CNN-C [52]. This network extracts $D = 256$ features from 128x128 grayscale image. The results for the PubFig83 [67] dataset with $T = 13813$ images and $C = 83$ classes are presented in Fig. 10 and Fig. 11.

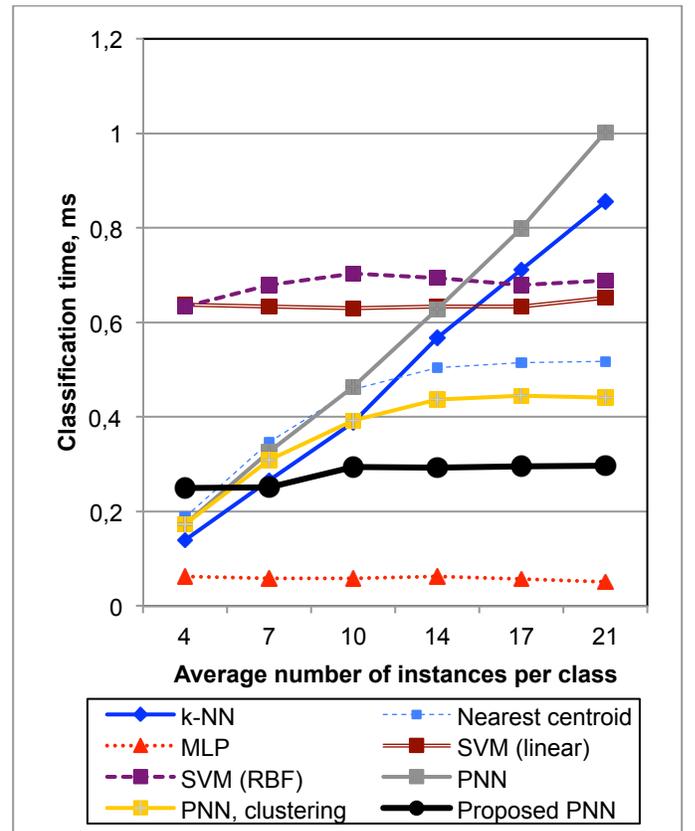

Fig. 11. Face recognition time (ms), PubFig83 dataset

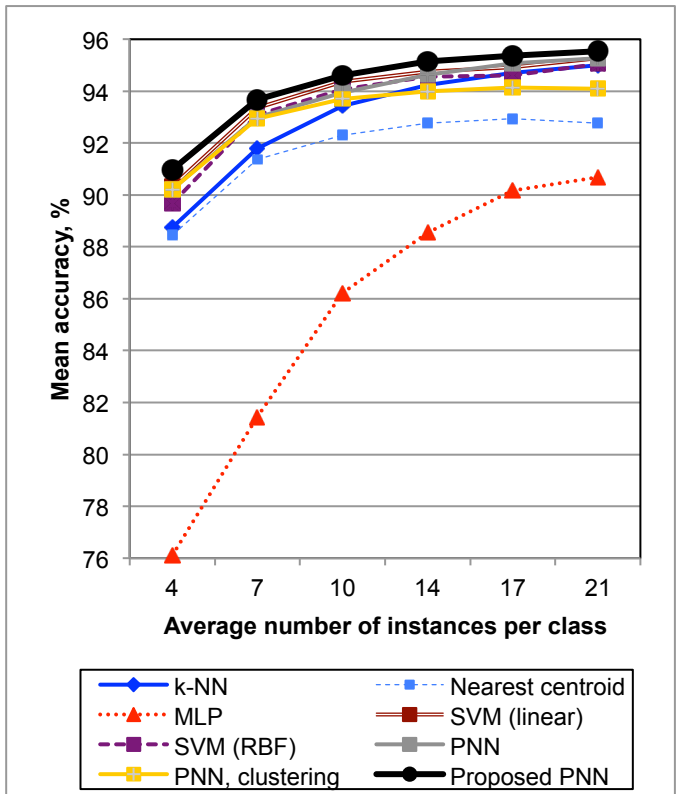

Fig. 10. Face recognition accuracy (%), PubFig83 dataset

The accuracies of all methods except the nearest centroid, the MLP and the PNN with clustering in this task are rather similar, though our modification is 0.5% more accurate when compared to the baseline PNN. At the same time, performance of our algorithm (Table I) gains of up to 4.5 times over the PNN and the k-NN. Our classifier is also 2.5-times faster than the SVM. Its time to recognize one face is even smaller that the average classification time of the reduced PNN with analysis of the clusters centroids.

In the last experiment we considered $T = 66000$ images from $C = 1000$ first classes of the CASIA-WebFace facial dataset [55]. It is one of the most difficult tasks in our study due to the large number of classes [12]. Fig. 12 and Fig. 13 present the estimates of the accuracy and recognition time, respectively.

In this task our network is considerably more accurate, than all other methods. Only MLP is characterized with similar accuracy if the training set is rather large. For example, our error rate is 1-4% lower than the error rate of the PNN. It is remarkable that conventional one-vs-all implementation of the SVM suffers from the large number of classes. Hence, our approach is 4.5% more accurate even when half of the images are stored in the training set. The gain in performance is even more noticeable. The usage of the exponential activations functions and the Fejér kernel speeds-up the PNN with the Gaussian Parzen kernel in 1.5-6 times. Despite the known reduction of the PNN, which decreases the accuracy at 2-3%, our approach still satisfies the statistical optimality property. Thus, we could say that the proposed modified network (Fig.

1) makes it possible to implement the image recognition with deep features much more efficiently, than the original PNN, and does not loose any of its advantages.

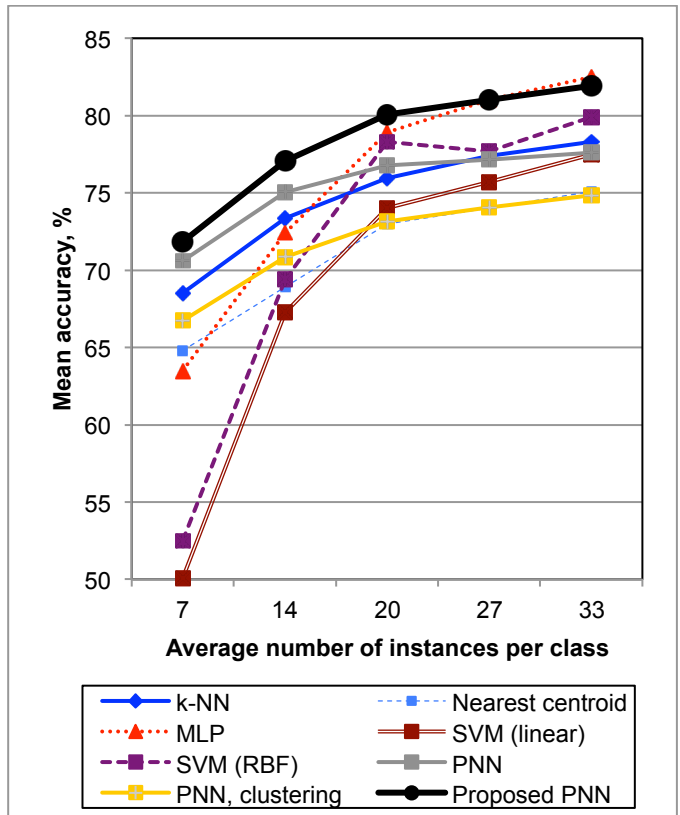

Fig. 12. Face recognition accuracy (%), CASIA-WebFace dataset

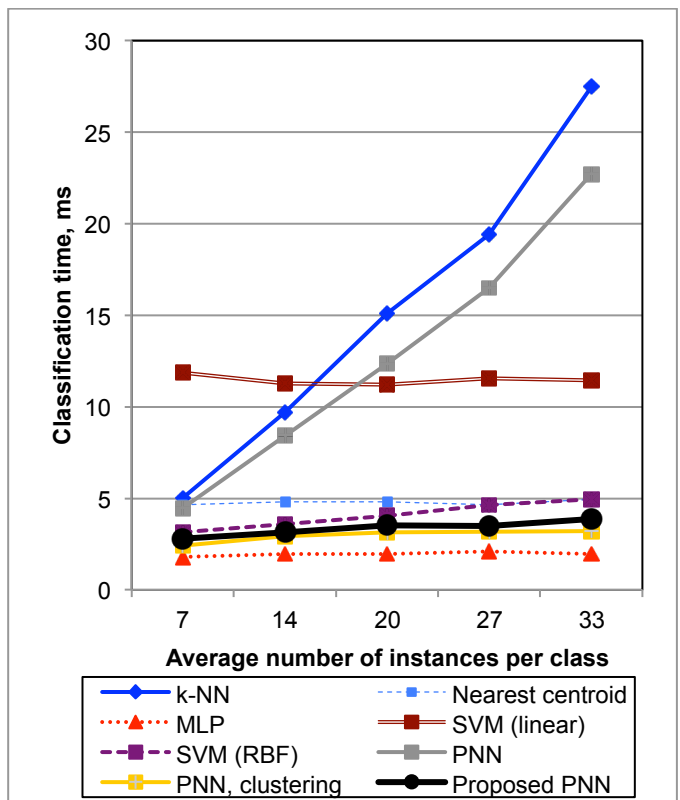

Fig. 13. Face recognition time (ms), CASIA-WebFace dataset

## V. CONCLUSION

In this paper we proposed a novel modification of the PNN (Fig. 1), which saves all advantages of the classical PNN including the fast training procedure and the convergence to the optimal Bayesian decision, but in theory the recognition performance and space complexity should be approximately $(R/C)^{2/3}$–times lower. These improvements are achieved by replacement of the exponential activation function in the pattern layer of the PNN to the *complex* exponential function (7). In the latter case it is possible to estimate the unknown probability densities using the simplified expressions of the trigonometric orthogonal expansions (10), (14) instead of the canonical kernel form (2), (9) with brute force of all training instances. Hence, the run-time and memory complexity become proportional to the cut-off parameter $J$, which is typically much less than the number of instances of each class. In the experimental study we describe a protocol for comparing image recognition methods using the well-known datasets (Caltech-101, Caltech-256, PubFig83, etc.) in the context of *small-sample-size* problem. Experiments with contemporary deep neural network models proved that our implementation of the MAP procedure (1) could be considered very promising in various image recognition tasks because of its high accuracy and rather low computational complexity. For sure, the proposed algorithm (Table I) is not the most accurate classifier in all cases, especially, if the number of instances per each class is rather large (Table V). However, it was shown that our approach makes it possible to deal with the drawbacks of the PNN caused by the brute force processing of all instances (1)–(3). Moreover, in contrast to the original PNN, our modification allows choosing the trade-off between accuracy and computational complexity (Fig. 7,8).

An important research direction is the application of our approach in other classification tasks [68], such as speech recognition, authorship attribution, etc. [18]. As the proposed classifier requires uncorrelated features from the range [-1; 1], it will be necessary to explore the normalization techniques for various features. Another important direction is the proper choice of the cut-off parameter (Fig. 7,8). The procedure of choosing the parameter $J$ as the median $J_d^*(c)$ (15) is appropriate only if the classes are balanced. It is also possible to investigate the implementation of the Bayesian networks [20] in our classifier in order to weaken the requirement of feature independence and evaluate the joint probability density functions only for several (2-3) dependent features. Finally, though our results for Caltech-101 dataset are rather inspiring, it is worth emphasizing that the fusion of classifiers [69] or simple stacking of descriptors for multi-scale features can improve the accuracy in 6-7% [7]. Hence, it is necessary to explore the usage of the proposed network in an ensemble of classifiers.

**Andrey V. Savchenko** was born in Gorky, USSR, in 1985. He received the M.S. degree in Applied Mathematics and Informatics in 2008 from Nizhny Novgorod State Technical University; PhD in Mathematical Modeling and Computer Science from State University Higher School of Economics (Moscow) in 2010; and Doctor of Science degree in system analysis and information processing from Nizhny Novgorod State Technical University.

Since 2008 he has been with the National Research University Higher School of Economics (Nizhny Novgorod, Russia), where he is currently a full professor in department of information systems and technologies. He is also a senior researcher in the Laboratory of Algorithms and Technologies for Network Analysis, National Research University Higher School of Economics. He is the author of one monograph and more than 50 articles. His research interests include statistical pattern recognition, image classification and biometrics.